\documentclass[runningheads]{llncs}
\usepackage{graphicx}
\usepackage{subfigure}
\usepackage{amsfonts}

\usepackage[utf8]{inputenc}
\usepackage{paralist}
\usepackage{capt-of}
\usepackage{booktabs}

\makeatletter
\newcommand{\printfnsymbol}[1]{%
  \textsuperscript{\@fnsymbol{#1}}%
}
\makeatother
\begin{document}
\title{Texture Retrieval in the Wild through Detection-based Attributes}
\author{Christian Joppi\inst{1}\thanks{These authors contributed equally to this work}  \and
Marco Godi\inst{1}\printfnsymbol{1} \and
Andrea Giachetti\inst{1} \and
Fabio Pellacini\inst{2} \and
Marco Cristani\inst{1}}
\authorrunning{C.Joppi et al.}
\institute{University of Verona, Verona, Italy \and Sapienza University of Rome}

\maketitle              % typeset the header of the contribution
\begin{abstract}
Capturing the essence of a textile image in a robust way is important to retrieve it in a large repository, especially if it has been acquired in the wild (by taking a photo of the textile of interest).
In this paper we show that a texel-based representation fits well with this task.
In particular, we refer to Texel-Att, a recent texel-based descriptor which has shown to capture fine grained variations of a texture, for retrieval purposes.
After a brief explanation of Texel-Att, we will show in our experiments that this descriptor is robust to distortions resulting from acquisitions in the wild by setting up an experiment in which textures from the \emph{ElBa} (an Element-Based texture dataset) are artificially distorted and then used to retrieve the original image. We compare our approach with existing descriptors using a simple ranking framework based on distance functions. Results show that even under extreme conditions (such as a down-sampling with a factor of 10), we perform better than alternative approaches.

\end{abstract}

%=================================INTRODUCTION=====================================================
\vspace{-0.3cm}
\section{Introduction}
\label{sec:intro}
\begin{figure}[t!]
\centering
\subfigure[]{\includegraphics[width=0.25\textwidth,height=3cm]{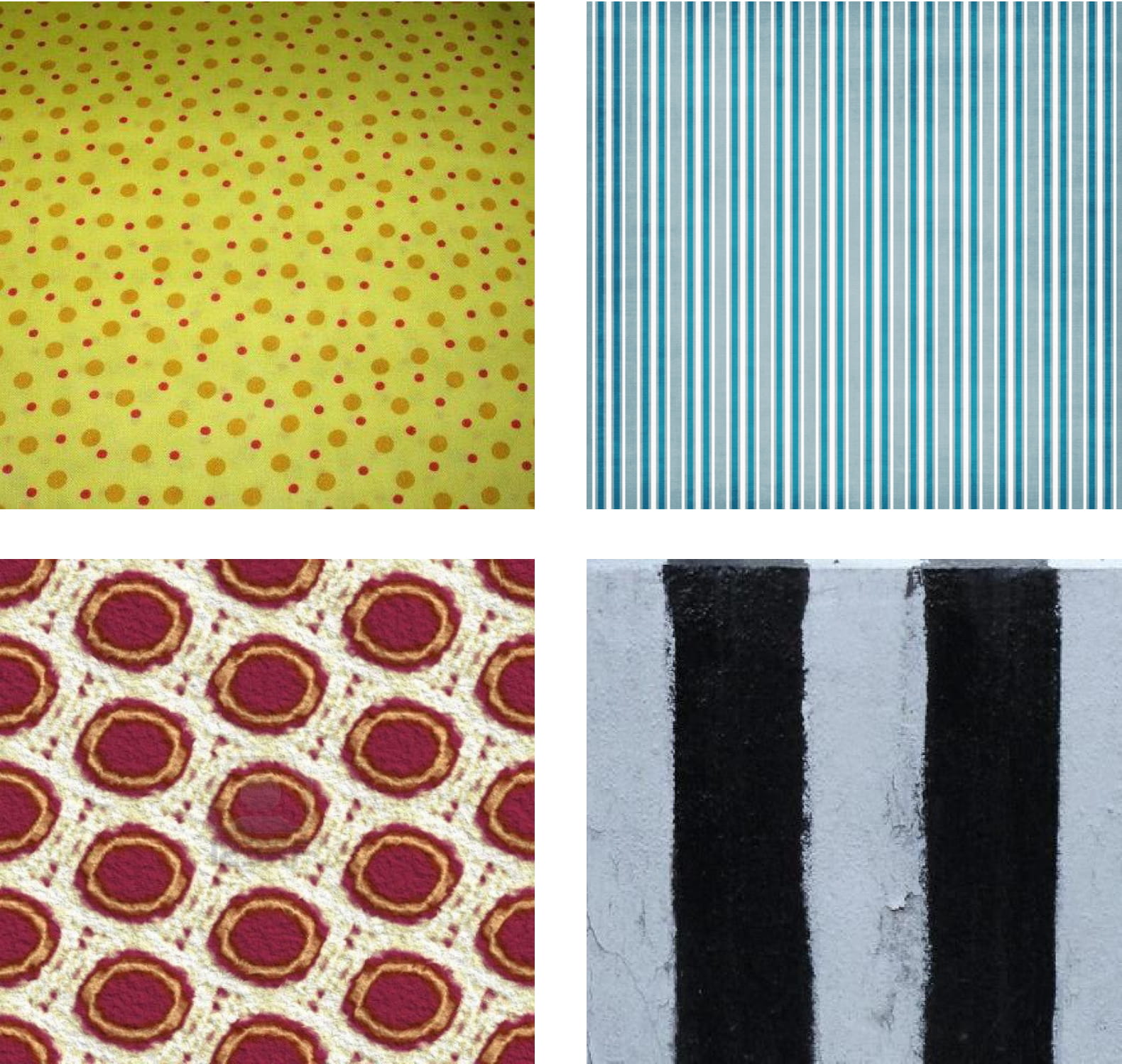}}
\hspace{0.2em}
\subfigure[]{\includegraphics[width=0.40\textwidth,height=3cm]{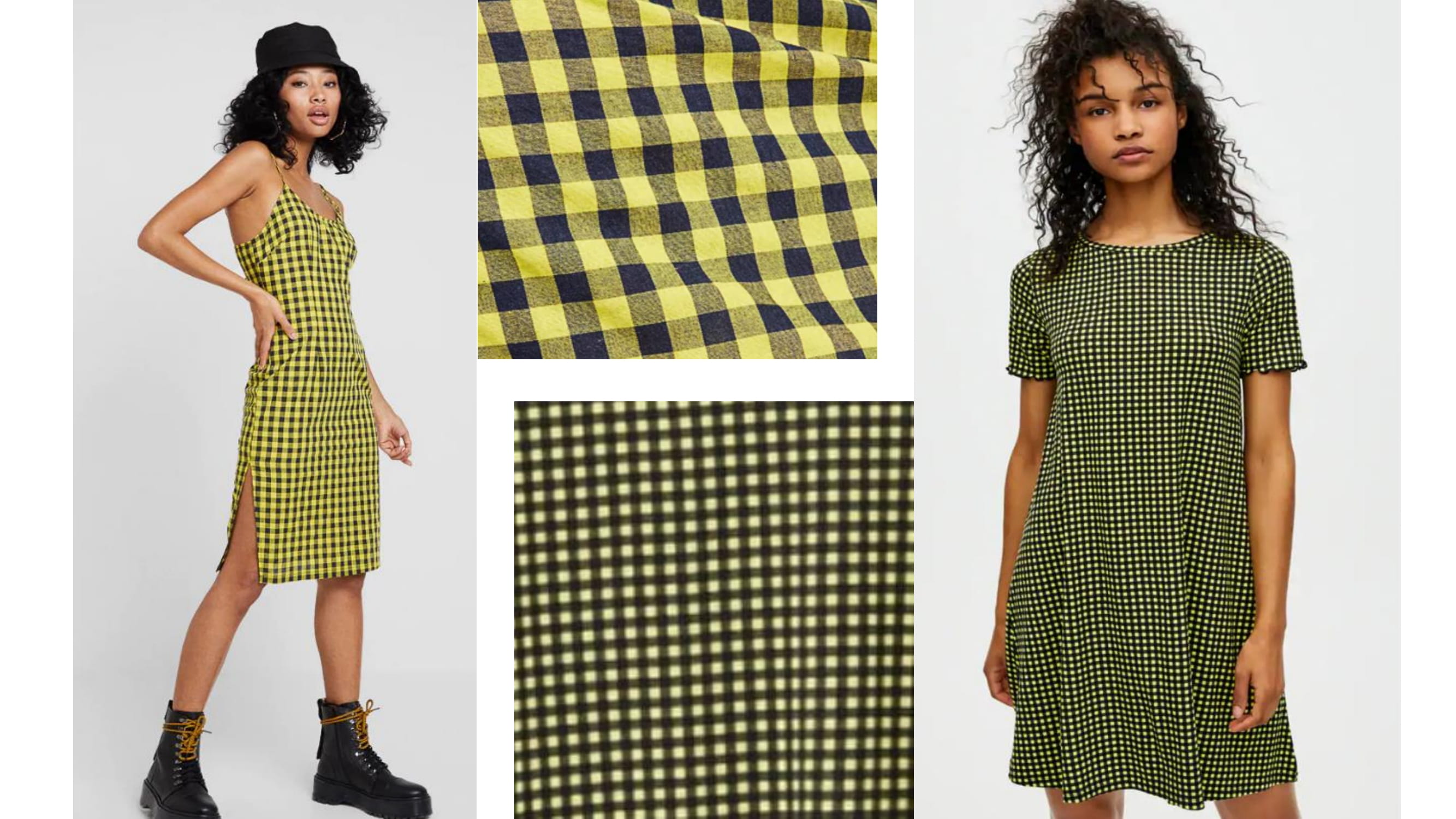}}
\hspace{0.2em}
\subfigure[]{\includegraphics[width=0.13\textwidth,height=3cm]{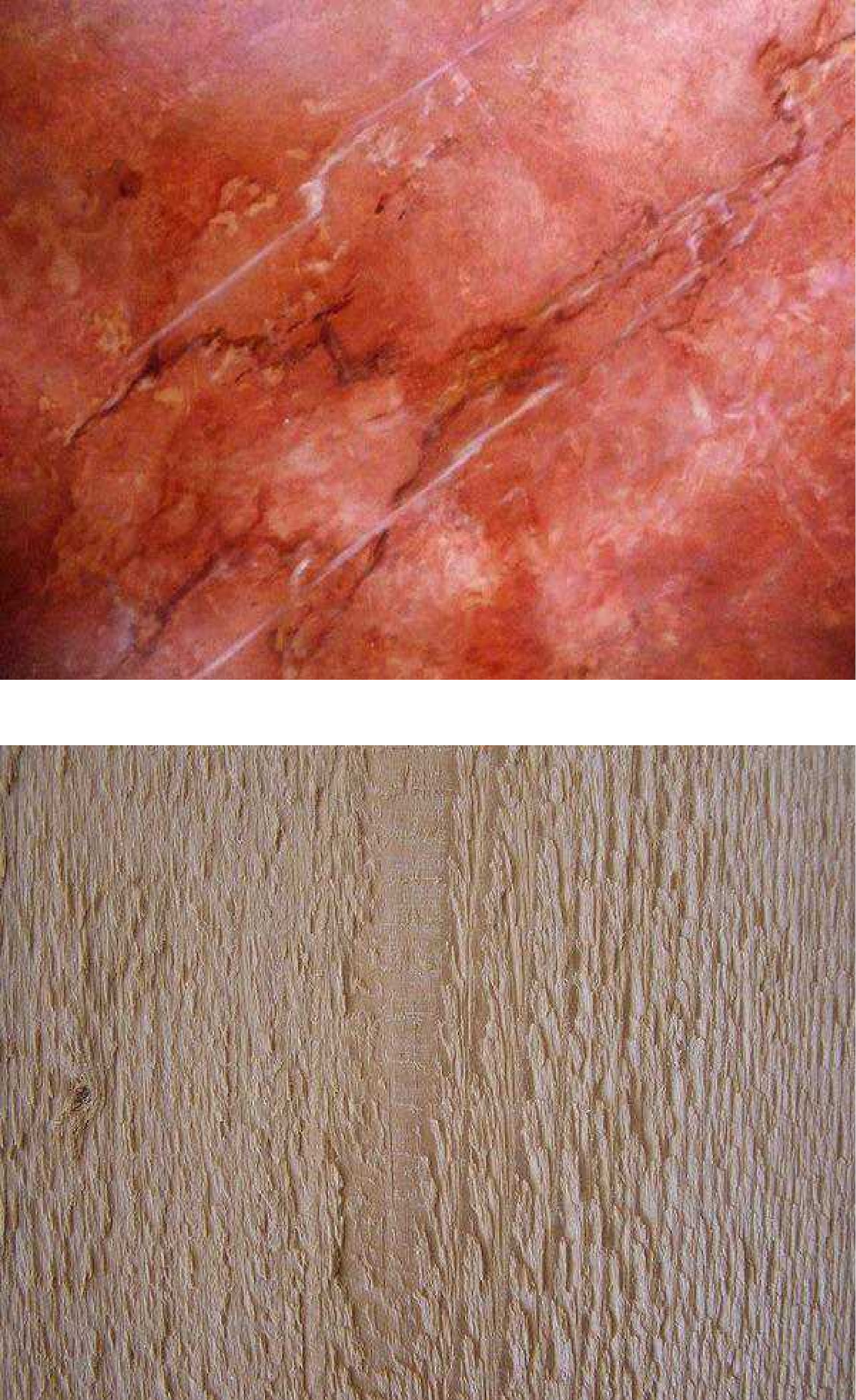}}
\hspace{0.2em}
\subfigure[]{\includegraphics[width=0.13\textwidth,height=3cm]{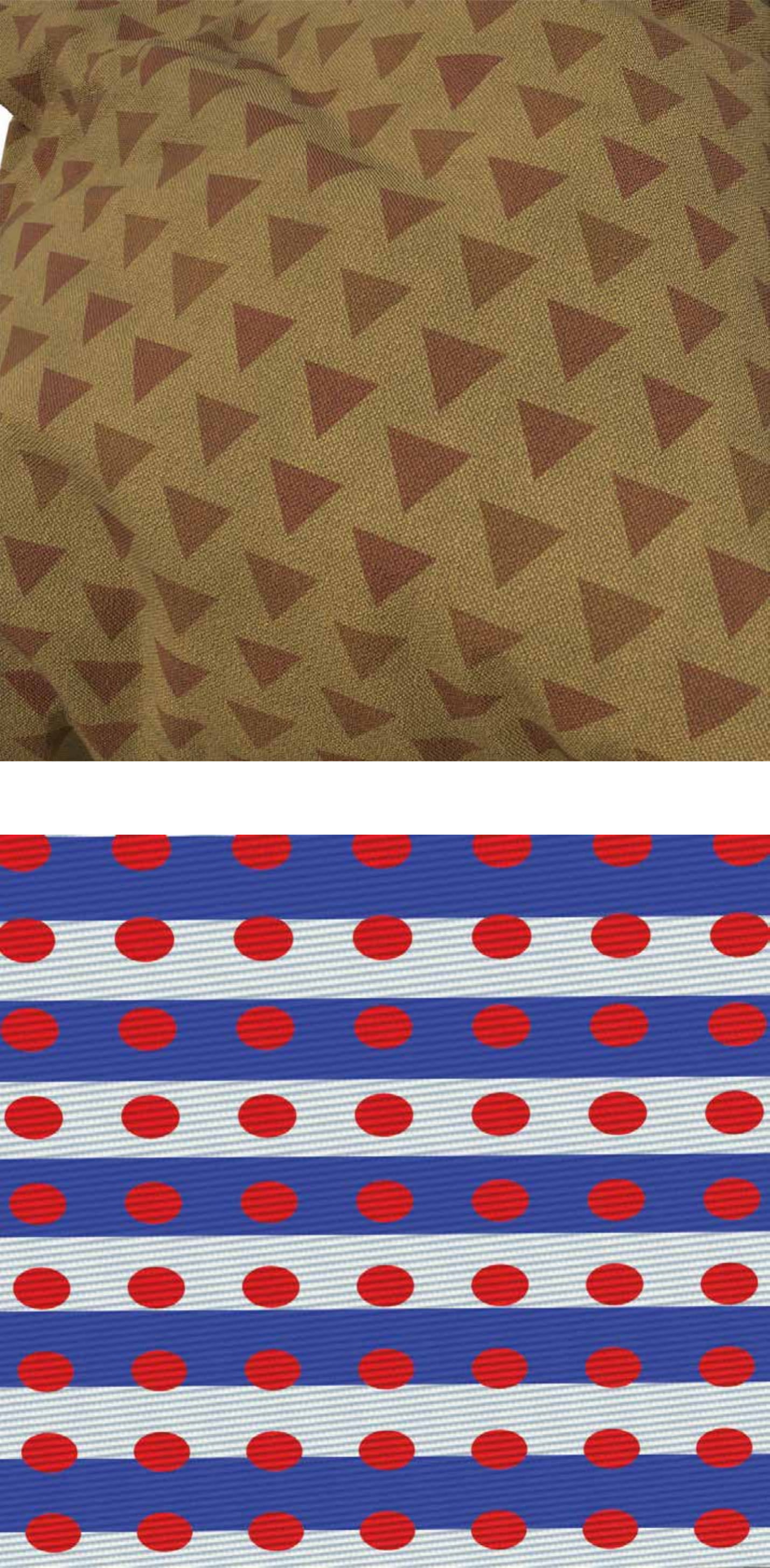}}
\caption{\small \label{fig:exte} (a) Examples of element-based textures in the DTD~\cite{cimpoi2014describing}: the \emph{dotted} (left) and \emph{banded} (right) classes are examples where texels are dots and bands, respectively; (b) Zalando shows for each clothing a particular on the texture; (c) examples of DTD~\cite{cimpoi2014describing} textures which are \emph{not} element-based: (\emph{marbled} on top and \emph{porous} on bottom); here is hard to find clearly nameable local entities; (d) examples of \emph{ElBa} textures: polygon on top, multi-class lined+circle texture on bottom.}
\vspace{-0.5cm}
\end{figure}

\vspace{-0.2cm}
\emph{Texels}~\cite{ahuja2007extracting} are nameable elements that, distributed according to statistical models (see Fig.~\ref{fig:exte}a-b), form textures that can be defined as \emph{Element-based}~\cite{ijiri2008example,ma2011discreet,ma2013dynamic,loi2017programmable}.
Textures of this kind are of interest in the textile, fashion and interior design industry, since websites or catalogues (containing many products) have to be browsed by users that want to buy or take inspiration from~\cite{kovashka2012whittlesearch,kovashka2015whittlesearch}. Two examples taken from the popular e-commerce website Zalando are shown in Fig.~\ref{fig:exte}b. For each item multiple pictures are usually available, including close-up pictures of the fabric highlighting the texture. Not all textures can be defined as Element-based; some can only be characterized at a \emph{micro} scale (\emph{e.g.} in the case of material textures in Fig.~\ref{fig:exte}c), but usually the patterns that decorate textile materials are based on repeated elements. 

In the fashion domain browsing for textures is a common task. 
A shopper that is in possession of an item (e.g. a shirt) with a specific pattern could wish to shop for another item (e.g. pants with a matching pattern) to combine with by taking a close-up picture to highlight the desired texture. 
A fashion designer could want to take inspiration from an existing garment with only a low resolution picture of the texture available.
In these scenarios, it would be useful to be able to search in a database for the desired texture using only a low-quality picture (i.e. in diverse lighting conditions and resolution) as a query. Texture retrieval that is robust to these conditions is an important addition for a fashion e-shop~\cite{pinterest2015visual,ebay2017visual} or for fashion designer tools~\cite{WhereToBuyItICCV15}.
To be able to achieve this for textures, it is very important to describe them and their structural information in an intuitive and interpretable way, in order to achieve a precise description that enables an accurate retrieval~\cite{smeulders2000content} based on the image content. 

For the purpose of achieving a discriminative and nameable description, attribute-based texture features~\cite{matthews2013enriching,roboticDatasetTex,cimpoi2014describing,surveytex2018} are explicitly suited.
In the literature, the 47 perceptually-driven attributes such as \emph{dotted, woven, lined,} etc. learned on the Describable Texture Dataset (DTD)~\cite{cimpoi2014describing} are the most known.
\\
These 47 attributes are limited in the sense that they describe the properties of a texture image \emph{as a single whole atomic entity}: in Fig.~\ref{fig:exte}a,
two different (element-based) attributes are considered: \emph{dotted} (left) and \emph{banded} (right) each one arranged in a column. Images in the same column, despite having the same attribute, are strongly different: for the dots, the difference is on the area; for the bands, the difference is on the thickness. 
In Fig.~\ref{fig:exte}b (Zalando examples), both garments come with the same ``checkered'' attribute, despite the different sized squares. 

It is evident that one needs to focus on the recognizable \emph{texels} that form textures to achieve a finer expressivity.
\\
\\
In this paper, we employ \emph{Texel-Att}~\cite{godi2019texelatt}, a fine-grained, attribute-based texture representation and classification framework for element-based textures. \\ 
The pipeline of Texel-Att first detects the single texels and describes them by using \emph{individual attributes}. Then, depending on the individual attributes, they are grouped and these groups of texels are described by \emph{layout attributes}. 

The Texel-Att description of the texture is formed by joining the individual and layout attributes, so that they can be used for classification and retrieval. The dimensionality of the Texel-Att descriptor isn't pre-defined, it depends on which attributes are selected for the task. In this paper, we just give some examples to illustrate the general framework.

A Mask-RCNN~\cite{he2017mask} is used to detect texels; this shows that current state-of-the-art detection architectures can produce element-based descriptions (further improvements are foreseeable as we will discuss later). We design \emph{ElBa}, the first \emph{El}ement-\emph{Ba}sed texture dataset, inspired by printing services and online catalogues\footnote{\url{https://www.spoonflower.com/}, \url{https://designyourfabric.ca/}, \url{https://patternizer.com/d0Wp} and \url{https://www.contrado.com/} respectively.}. By varying in a \emph{continuous} way element shapes and colors and their distribution, we generate realistic renderings of 30K texture images in a procedural way using a total of 3M localized texels. Layout attributes such as local symmetry, stationarity and density are known by construction.
\\

In the experiments we show that, using the attribute-based descriptor that we extract with our framework, we are able to retrieve textures in a more accurate way under simulated  image conditions mimicking real-world scenarios. The performance of our approach is compared against state of the art texture descriptors of different kinds to show the usefulness of our approach.

We also show qualitative results to highlight the steps of the employed framework, such as the texel detection (detailed in Sec~\ref{sec:textiler}). 

%=================================FRAMEWORK=====================================================
\vspace{-0.2cm}
\section{Method}
\vspace{-0.2cm}
In this section we explain the Texel-Att framework step-by-step. Then we propose a simple method for texture retrieval that can be employed with this framework.
\vspace{-0.2cm}
\subsection{The Texel-Att Framework}\label{sec:textiler}
\vspace{-0.2cm}

\begin{figure}[t!]
\centering
   \includegraphics[width=\linewidth]{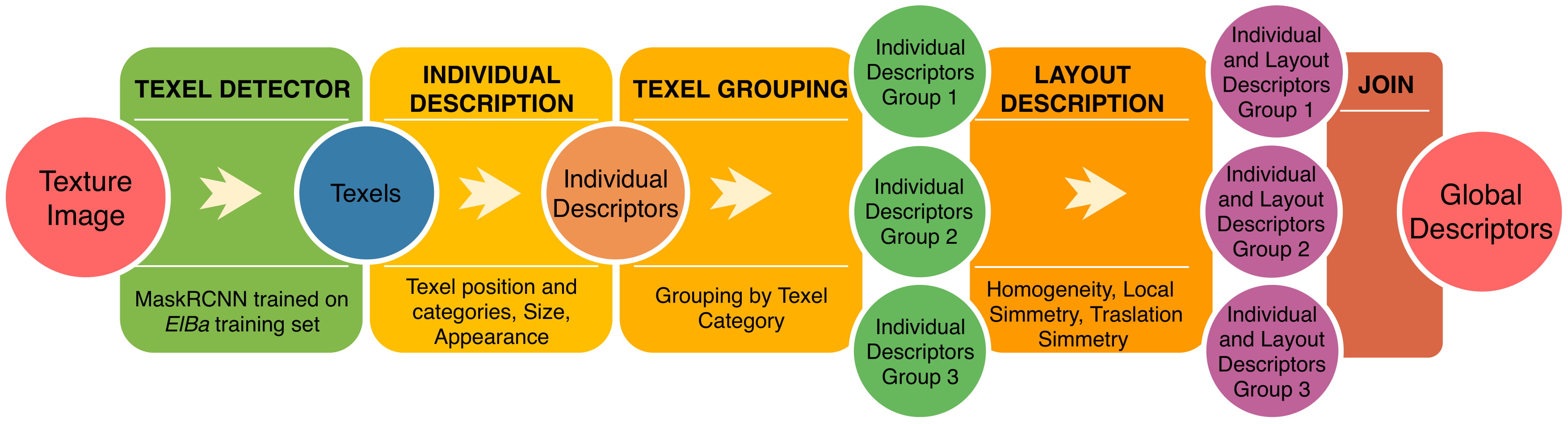} 
    \caption{\small \label{fig:scheme} Block diagram of the formation of the Texel-Att element-based texture descriptor. On the bottom of each plate, the specific choices made in this paper, which can be varied.}
    \vspace{-0.5cm}
\end{figure}

Fig. \ref{fig:scheme} shows a block diagram of the Texel-Att description creation pipeline. 

The main concept is extracting texels using an object detection framework (trained for the task). Then, texels are described with \emph{individual} attributes, \emph{i.e.} labelled according to category, appearance and size. Texels are then grouped and filtered according to the individual labels. For each group, descriptions of the spatial layout of groups are estimated and aggregated into \emph{layout} attributes. The composite Texel-Att descriptor is formed by individual and layout attributes.  
In the following, each processing block is detailed.
 
\textbf{Texel Detector}. The Mask-RCNN~\cite{he2017mask} model handles the texel detection by localizing (with bounding boxes and segmentation masks) and classifying objects. 
The model is trained on the \emph{ElBa} dataset's training set, learning to detect and classify texels such as \emph{lines}, \emph{circles}, \emph{polygons} (see Sec.~\ref{sec:ElBa_dataset}). Texels are easily handled in any displacement (while a few years ago it was a quite complicated task limited to specific scenarios \emph{i.e.}, lattices~\cite{gui2011texel,liu2015patchmatch}).

\textbf{Individual description of texels.} By using attributes related to shape and human perception it is possible to characterize each detected texel; in particulare we make use of: (i) the \textit{label} indicating its shape, classified by the Mask-RCNN model; (ii) histogram of 11 \textit{color}s using a color naming procedure~\cite{van2009learning}; (iii) \textit{orientation} of texels; (iv) \textit{size} of texels, represented by the area in pixels.
By aggregating (e.g. through averages or histograms, see in the following sections) it is possible to characterize the whole texture. It is worth noting that in this work we are not showing ``the best'' set of features, but we are highlighting the portability and effectiveness of the framework; in fact, different attributes could be used instead.

\textbf{Texel Grouping}
Texels with the same appearance are clustered, so that spatial characteristics of similar elements can be captured using layout attributes. In this work we simply group texels by the assigned shape labels (\emph{circle}, \emph{line} or \emph{polygon}). Groups with less than 10 texels are removed. 

\textbf{Layout description of texels.}
Spatial characteristics of each texel group, are described by measuring attributes using the spatial distribution of the centroids of the texels. We can refer to the literature on spatial points pattern analysis, where measures for symmetry, randomness, and regularity ~\cite{diggle1983statistical,velazquez2016evaluation,baddeley2015spatial} are available; we select a simple and general set of measures. They are:
(i) texel \textit{density}, \emph{e.g.} the average number of texels per unit of area (for circles and polygons) or line density (\emph{e.g.} by projecting centroid on to the direction perpendicular to their principal orientation density is measured on one spatial dimension).
(ii)  Quadratic counts-based \textit{homogeneity} evaluation~\cite{illian2008statistical}: the original image is divided into a number of patches and a $\chi^2$ test is performed to evaluate the hypothesis of average point density in each patch. Similarly to the previous case, we estimated a similar 1D feature on the projection for lines.
(iii) Point pair statistics~\cite{zhao2011translation}: the histogram of \textit{vectors orientation} is estimated using point pair vectors for all the texel centers.
(iv)  \textit{Local symmetry}: we considered the centroids' grid for circles and polygons and measured, for 4-points neighborhoods of points, the average reflective
self-similarity after their reflection around the central point. The average point distance is used as a distance function. Neighborhood size is used to normalize it.
\textit{Translational symmetry} is estimated in a similar way by considering 4-point neighborhoods of the centroids traslated by the vectors defined by point pairs in the neighborhood and measuring the average minimum distance of those points. For line texels, we compute on 1D projections.

We report the dimensionalities for each of these attributes in Tab.~\ref{tab:descriptorSize}. Multi-dimensional attributes are histograms, while 1-dimensional ones are averages.
By concatenating and Z-normalizing spatial pattern attributes, individual texel attributes statistics and the color attributes of the \textit{background}, the final descriptor for the texture is built.

\begin{table}[t]
\small
\begin{flushleft}
\begin{minipage}{.35\textwidth}
\scalebox{0.6}{
\begin{tabular}{ccccc}
\toprule
Label & Color & Orientation & Size & \textbf{Total}\\
Histogram &  & Histogram &  & \\
\toprule
3 & 11 & 3 & 1 & 18\\
\bottomrule
\end{tabular}
}
\end{minipage}
\begin{minipage}{.60\textwidth}
\scalebox{0.67}{
\begin{tabular}{ccccccc}
\toprule
Density & Homogeneity & Vector  & Local  & Traslational  & Background  & \textbf{Total}\\
 &  &  Orientations &  Symmetry &  Symmetry &  Color & \\
\toprule
1 & 1 & 3 & 1 & 1 & 11 & 18\\
\bottomrule
\end{tabular}
}
\end{minipage}

\end{flushleft}
\small
\caption{\small \label{tab:descriptorSize} Dimensionality of descriptor attributes. On the left, the attributes computed from the individual characterization of texels; on the right, attributes computed from statistics resulting from the spatial layout. The total dimensionality of the descriptor is 36. }
\vspace{-0.5cm}
\end{table}
\vspace{-0.2cm}
\subsection{Element-Based Texture Retrieval}\label{sec:retrieval}
\vspace{-0.2cm}
The descriptor detailed in the previous section can be used to compute distances between element-based textures using the corresponding attributes. We define \emph{database set} the set of images that we want to search into using a \emph{query image}. The idea is that database texture closest to the query image (in terms of descriptor distance) are also the most similar ones in the database set.

The pipeline is as follows: a query image (e.g. a picture of a textured captured by a user) is processed by the Texel detector, allowing for the computation of individual and layout attributes and thus obtaining a descriptor. A standard distance function (such as cosine distance) is computed between every database image and the query image. The database set is then sorted according to the distance and the resulting ranking can be shown to the user for browsing.

%=================================DATASET=====================================================
\vspace{-0.2cm}
\section{\emph{ElBa}: Element-Based Texture Dataset}
\label{sec:ElBa_dataset}
\vspace{-0.2cm}
While available datasets such as the DTD~\cite{cimpoi2014describing} include some examples of element-based textures mixed with other texture types (Fig.~\ref{fig:exte}(a)), there is no dataset focused on this particular domain. In this work, we present \emph{ElBa}, the first element-based texture dataset.
As shown in Fig.~\ref{fig:exte}(d), photo-realistic images are included in the \emph{ElBa} dataset. Training a model with synthetic data is a common practice~\cite{tremblay2018training,barbosa2018looking} and annotations for texels are easily made available as an output of the image generation process. \emph{Layout} attributes and \emph{individual} ones (addressing the single texel can be varied in our proposed parametric synthesis model. For example individual attributes such as texel shape, size and orientation and color can be varied. 
Available shapes are \emph{polygons} (squares, triangles, rectangles), \emph{lines} and \emph{circles} (inspired by the 2D shape ontology of~\cite{niknam2011modeling}). The idea is that these kind of shapes are common in geometric textiles and they approximate other more complex shapes.
Orientation and size are varied within a range of values. We choose colors from color palettes to simulate a real-world use of colors.

\begin{figure}[t!]
\centering
\includegraphics[width=0.85\textwidth]{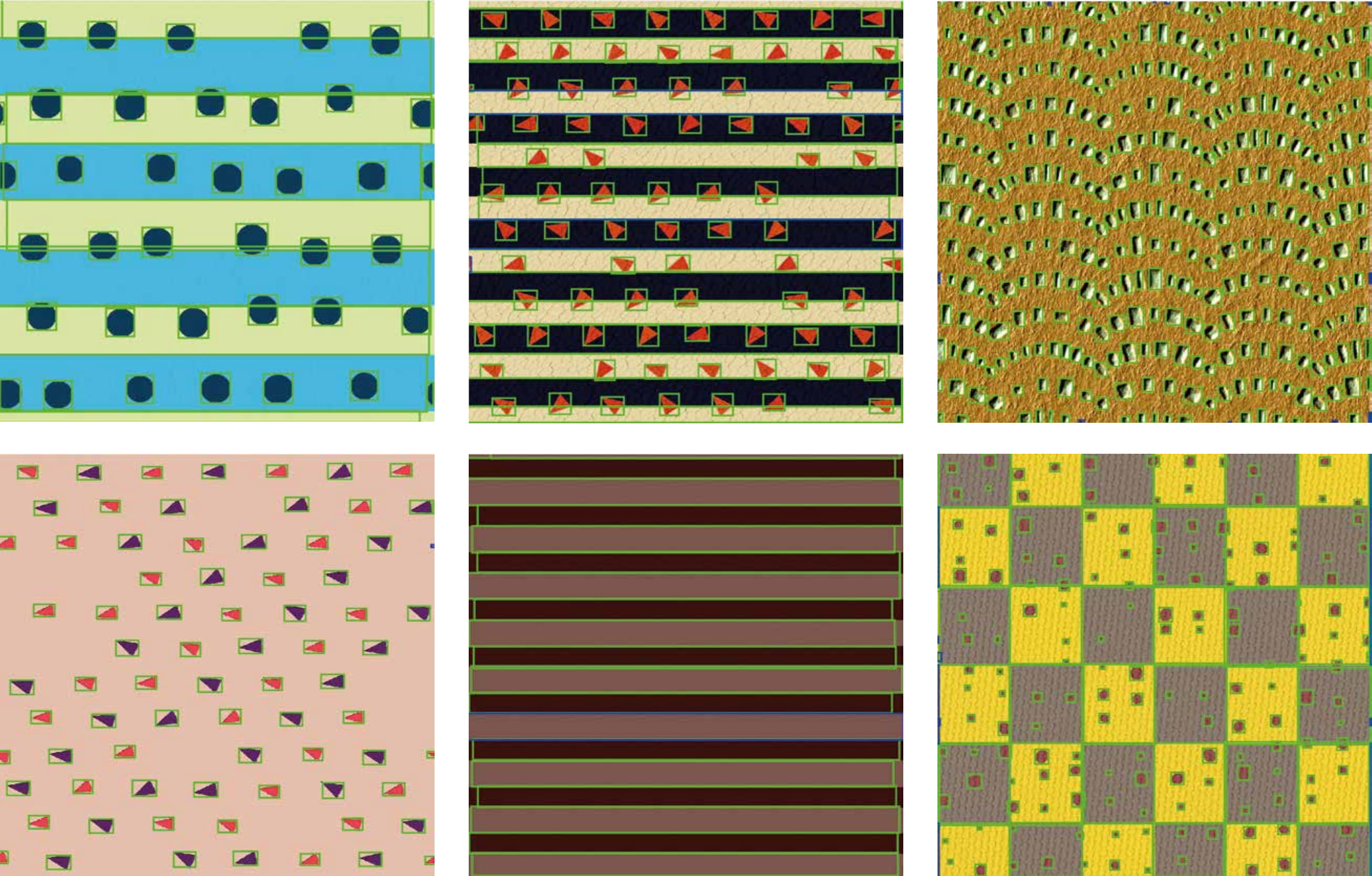}
\caption{\small \label{fig:detRes} Texel-Att detection qualitative results on \emph{ElBa} datasets. In green the correct detections, in red the false positives and in blue the false negatives.}
\vspace{-0.5cm}
\end{figure}

As for Layout Attributes, we select different 2D layouts based on symmetries to place texels. Linear and grid-based layouts are considered; one or two non-orthonormal vectors define the translation between texels in the plane. With this parametrization, we can represent several tilings of the plane. As for randomized distributions, we jitter the regular grid, creating a continuous distribution between randomized and regular layouts.

We also consider multiple element shapes within a single image, creating for example dotted+striped patterns. Each group of elements of the same shape is distributed with its own spatial layout, creating arbitrary multi-class element textures as in Fig.~\ref{fig:exte}(d).

We made use of Substance Designer for pattern generation (which gives high-quality output and pattern synthesis, and is easily controllable) and IRay ( which is a physically-based renderer) \footnote{\url{https://www.allegorithmic.com/} and \url{https://bit.ly/2Hz4ZVI} respectively.}.
Substance gives high-quality pattern synthesis, easy control and high-quality output including pattern antialiasing.
Low-frequency distortions of the surface of the plane where the pattern is represented and high frequency patterns are added to simulate realistic materials.

A total of 30K texture images (for a total 3M annotated texels) rendered at a resolution of $1024 \times 1024$ has been generated by this procedure. For each image ground-truth data (such as texel masks, texel bounding boxes and attributes) is available.
\emph{ElBa} does not come with a partition into classes: differently from other datasets used in texture analysis semantic labels for classification tasks can be computed from ground truth attributes or by user studies.

The dataset is randomly partitioned with a 90/10 split for, respectively, training and testing set.

%=================================EXPERIMENTS=====================================================
\begin{table}[t!]
\setlength{\tabcolsep}{1.3em}
\begin{center}
\scalebox{0.90}{
\begin{tabular}{cccc}
\toprule
\textbf{Distortions} &\textbf{Tamura}~\cite{tamura1978textural} & \textbf{FV-CNN}~\cite{cimpoi2016deep}   & \textbf{Texel-Att}\\
\bottomrule
\bottomrule
Down-Sampling (100x100) and\\Impulsive Noise (p=0.2) & 0.1380 & 0.3304 & \textbf{0.6618}\\
\bottomrule
Down-Sampling (200x200) and\\Impulsive Noise (p=0.2) & 0.2103 & 0.4811 & \textbf{0.8011}\\
\bottomrule
Down-Sampling (300x300) and\\Impulsive Noise (p=0.2) & 0.2284 & 0.5640 & \textbf{0.8560}\\
\bottomrule
Down-Sampling (100x100) and\\Radial Lighting Effect& 0.1611 & 0.4394 & \textbf{0.6356}\\ 
\bottomrule
Down-Sampling (200x200) and\\Radial Lighting Effect& 0.1728 & 0.8001 & \textbf{0.8746}\\ 
\bottomrule
Down-Sampling (300x300) and\\Radial Lighting Effect& 0.2708 & 0.8855 & \textbf{0.9376}\\
\bottomrule
\end{tabular}}
\end{center}
\caption{\small \label{tab:RetriRes} \emph{AUC (Area Under Curve)} for each distortion variant. Texel-Att performs better on every one of them. The related CMC are shown in Fig.~\ref{fig:Retr_plot}.}
\vspace{-0.7cm}
\end{table}

\vspace{-0.2cm}
\section{Experiments}\label{sec:exp}
\vspace{-0.2cm}
Experiments show the potential of our framework for the description of element-based textures, with a focus on difficult environmental conditions (low resolution and diverse lighting) ensuring an accurate retrieval inside large catalogues of textures in real-world applications.
\vspace{-0.2cm}
\subsection{Qualitative Detection Results}\label{sec:exp:detection}
\vspace{-0.2cm}
 We briefly show the detection results over our dataset, a fundamental step of our framework, through some qualitative results in Fig~\ref{fig:detRes}. Texels are highlighted by bounding boxes which are then used to compute the attributes (described in Sec.~\ref{sec:textiler}) that we employ in the following experiment.
\vspace{-0.2cm}
 \subsection{Texture Retrieval Results}\label{sec:exp:retrieval}
 \vspace{-0.2cm}
 In this experiment, we highlight the effectiveness of Texel-Att in a retrieval task under simulated real-world conditions following the procedure detailed in Sec.~\ref{sec:retrieval}. We compare our approach with both state-of-the-art texture descriptor FV-CNN~\cite{cimpoi2014describing} and Tamura attribute-based descriptor~\cite{tamura1978textural}. The \emph{database set} for this retrieval experiment is the whole test partition of the \emph{ElBa} dataset (composed of $\sim$3000 images). To simulate the real challenging conditions, we generated 6 variants of each image, down-sampling at one of 3 different resolutions (100x100, 200x200, 300x300) and up-sampling them back to the original image size (1024x1024).
 Then we apply one of the following distortions:
\begin{itemize}
    \item impulsive noise with a pixel's probability of 0.2 over all the image;
    \item radial lighting effect, increasing the brightness on a random point on the image and gradually decreasing it more in each pixel the farther from the chosen point it is.
\end{itemize} 

\begin{figure}[t!]
    \begin{center}
        \includegraphics[width=0.7\linewidth]{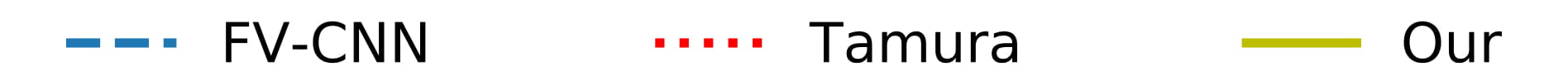}
    \end{center}
    \subfigure[]{\includegraphics[width=0.31\linewidth]{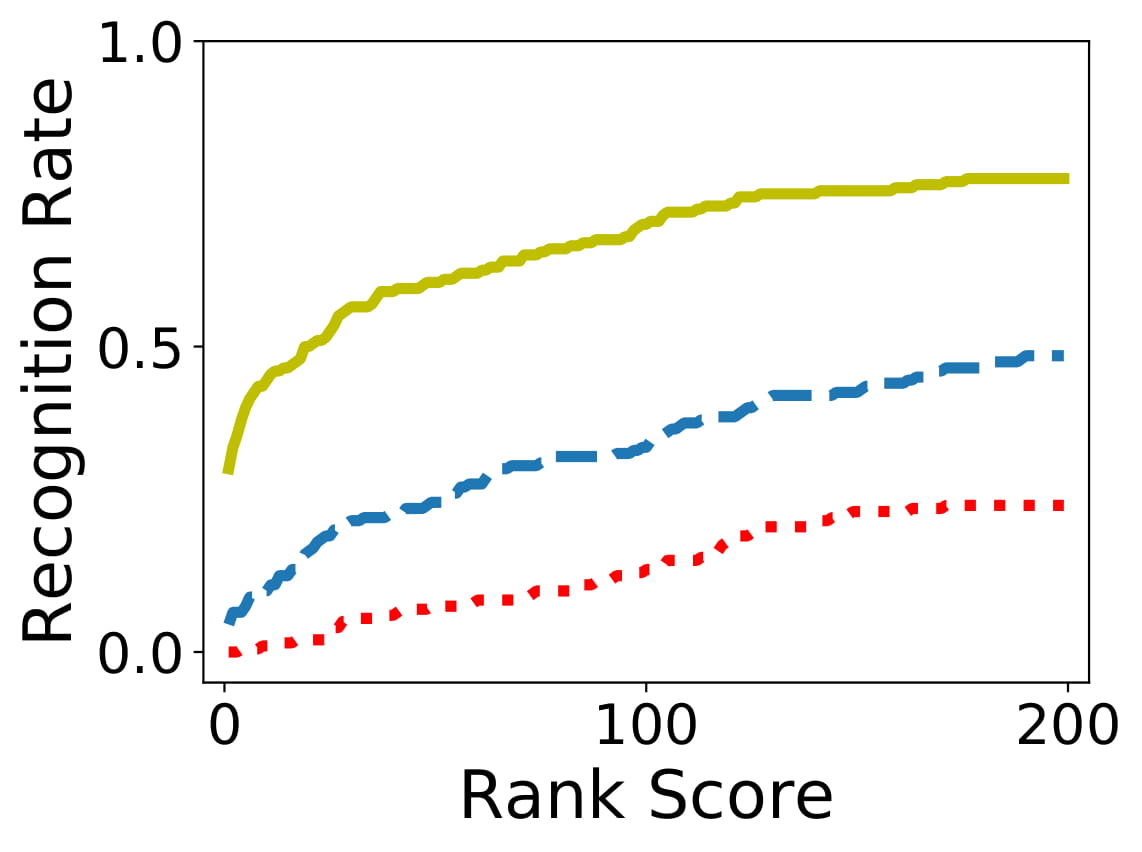}}
    \hspace{0.02\linewidth}
    \subfigure[]{\includegraphics[width=0.31\linewidth]{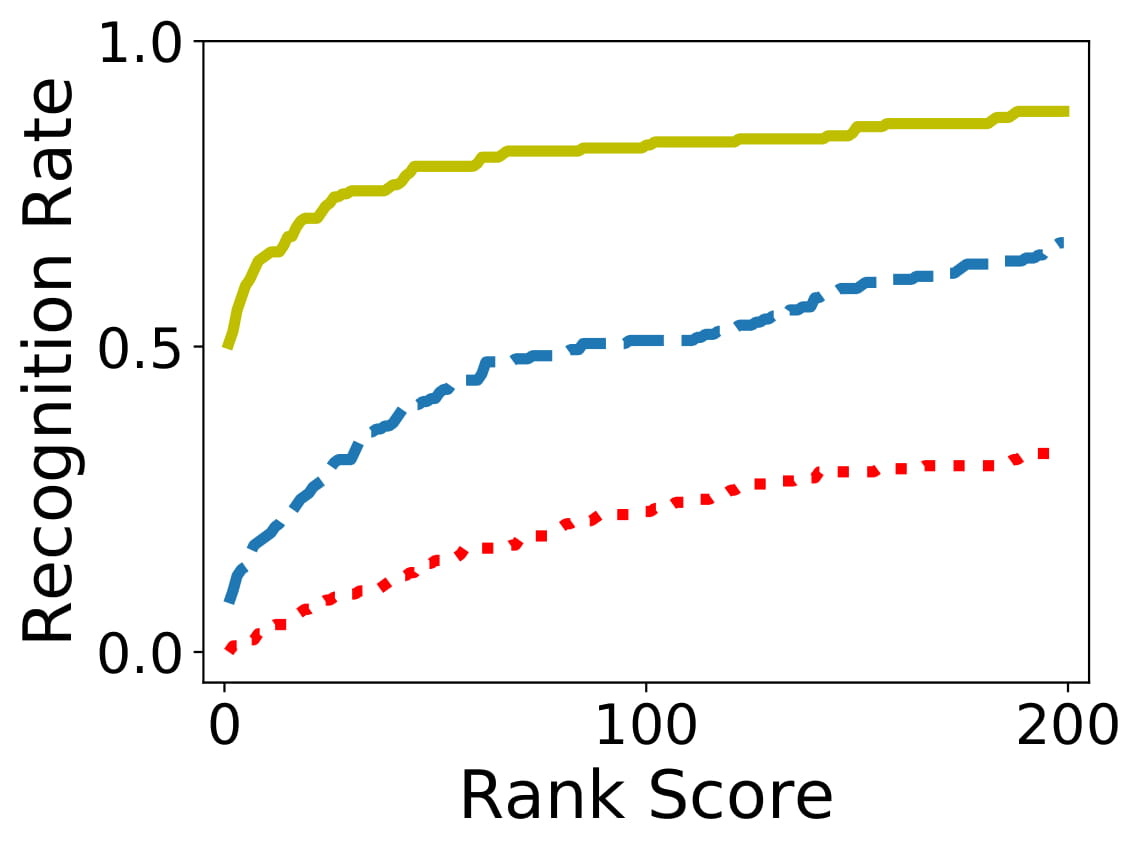}}
    \hspace{0.02\linewidth}
    \subfigure[]{\includegraphics[width=0.31\linewidth]{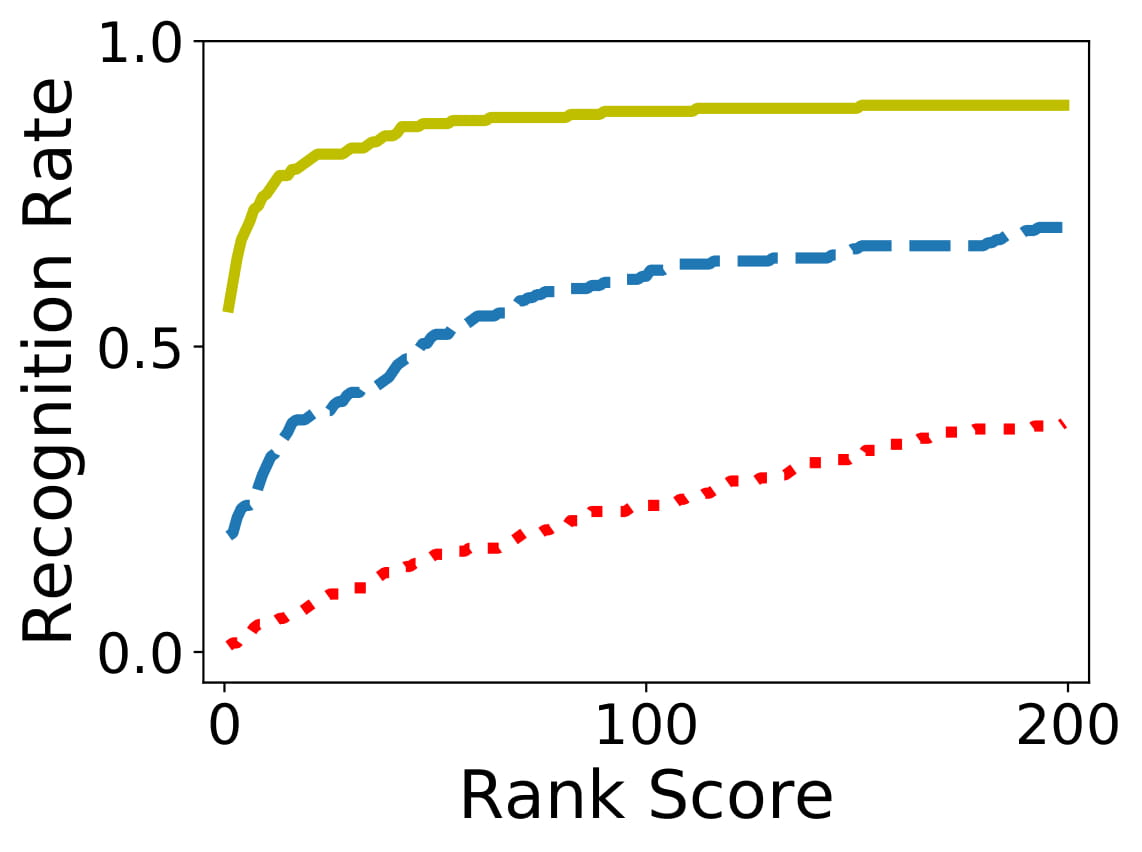}}
    
    \subfigure[]{\includegraphics[width=0.31\linewidth]{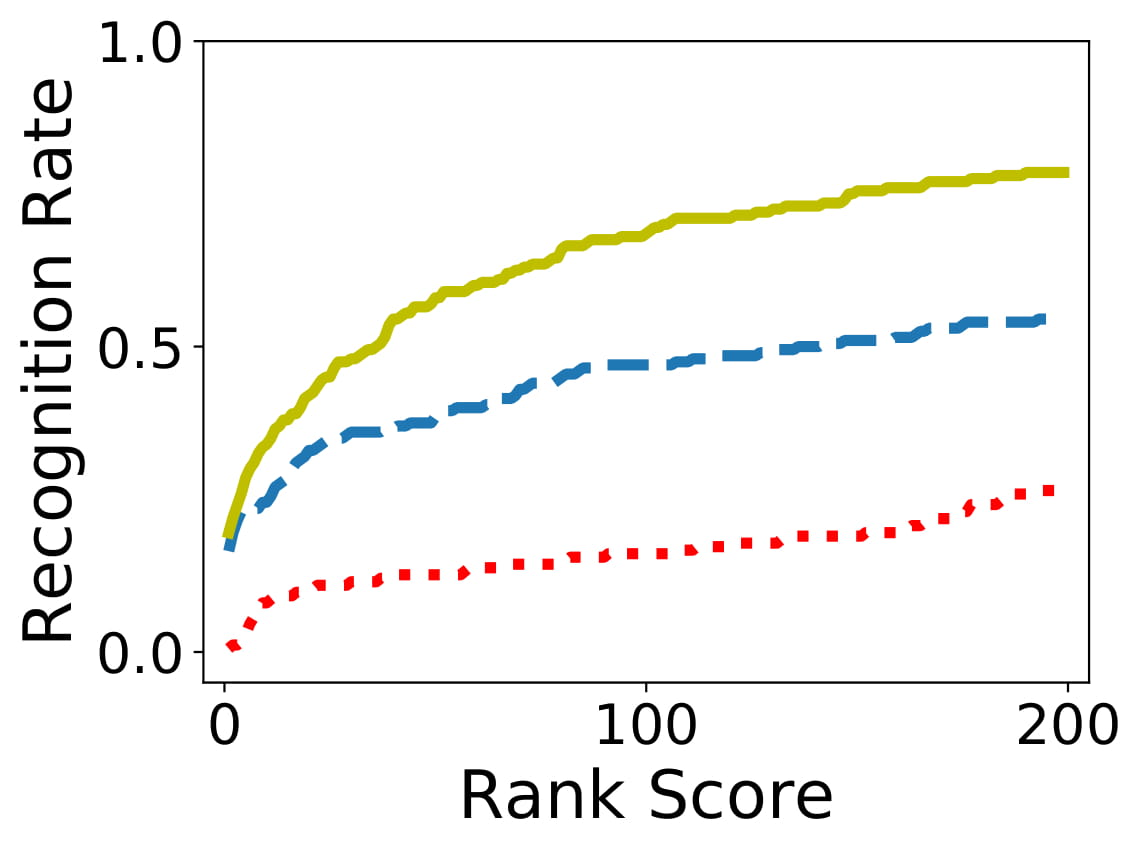}}
    \hspace{0.02\linewidth}
    \subfigure[]{\includegraphics[width=0.31\linewidth]{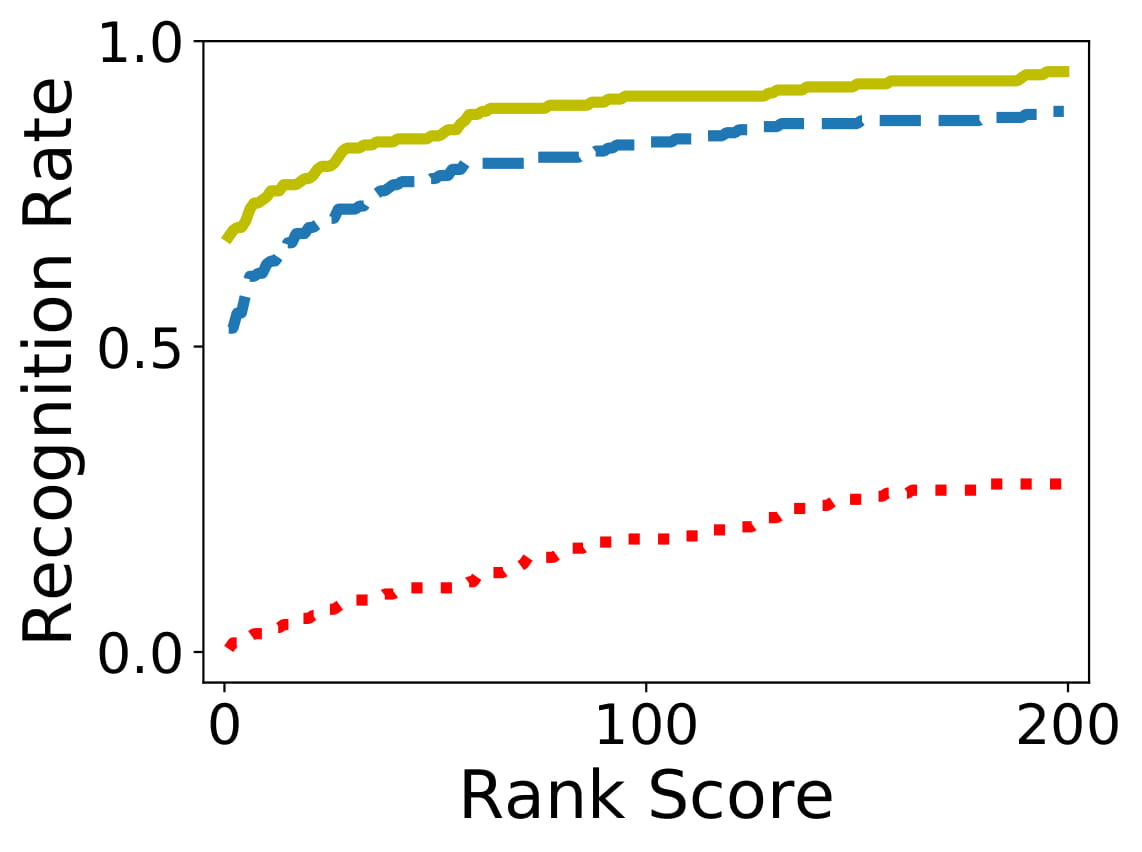}}
    \hspace{0.02\linewidth}
    \subfigure[]{\includegraphics[width=0.31\linewidth]{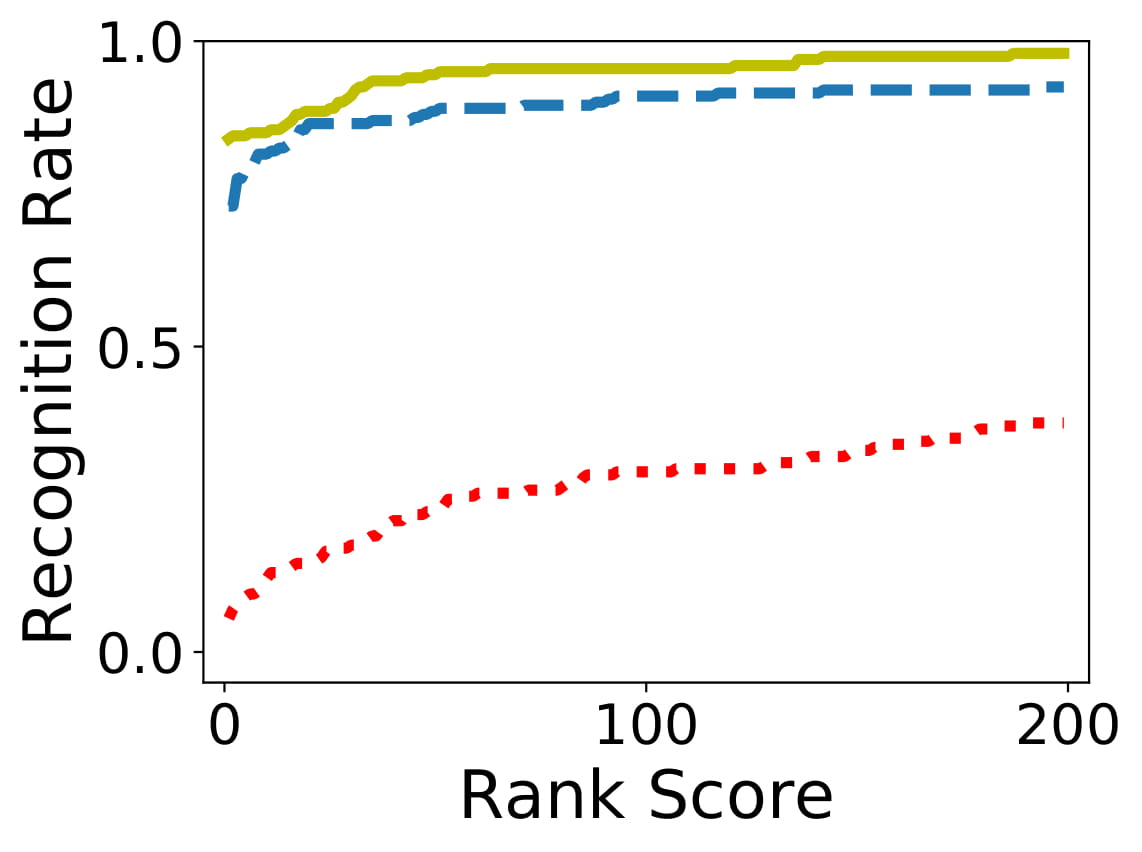}}
    \caption{\small  \label{fig:Retr_plot} \emph{CMC curves} on the retrieval experiments. Different plot for different variants of distortion: (a) 100x100 down-sampling and impulsive noise (b) 200x200 down-sampling and impulsive noise (c) 300x300 down-sampling and impulsive noise (d) 100x100 down-sampling and radial lighting effect. (e) 200x200 down-sampling and radial lighting effect. (f) 300x300 down-sampling and radial lighting effect. On the x axis the rank score (first 200 positions). On the y axis the recognition rate.}
    \vspace{-0.5cm}
\end{figure}

\begin{figure}[ht!]
\centering
\includegraphics[width=\textwidth]{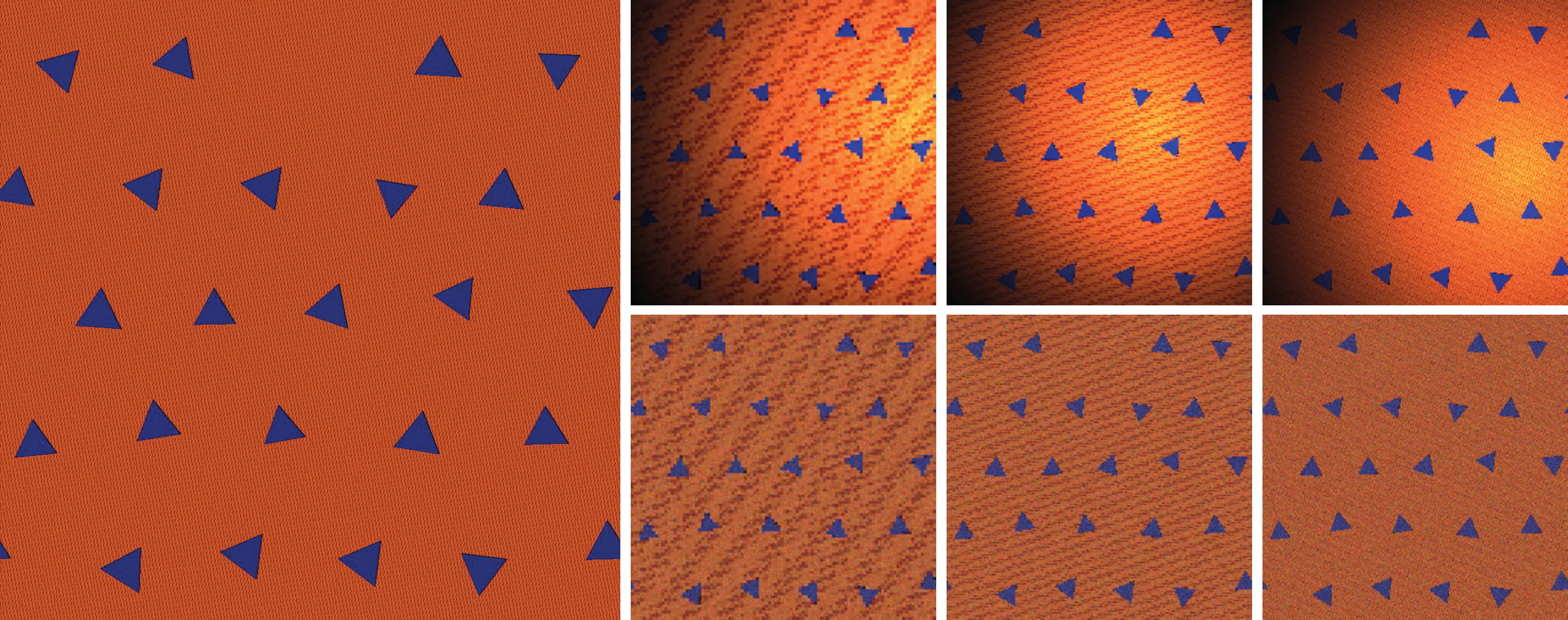}
\includegraphics[width=\textwidth]{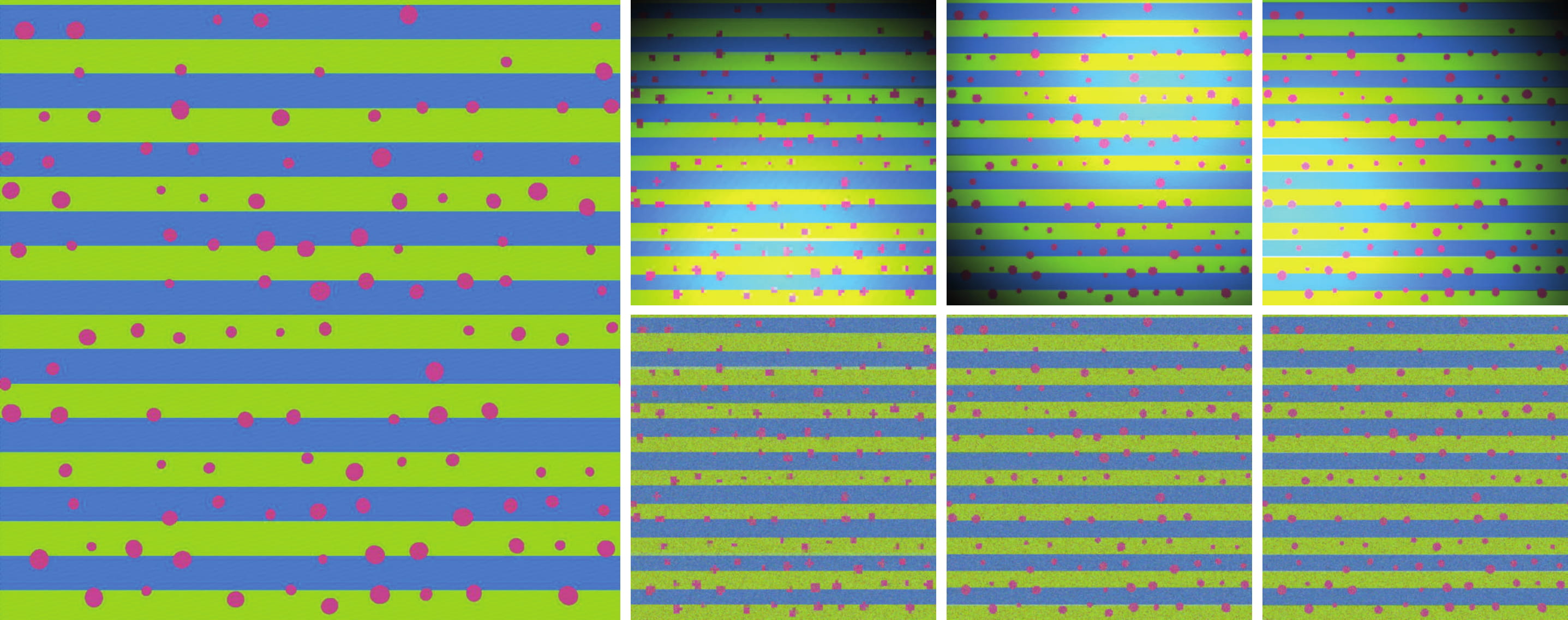}
\includegraphics[width=\textwidth]{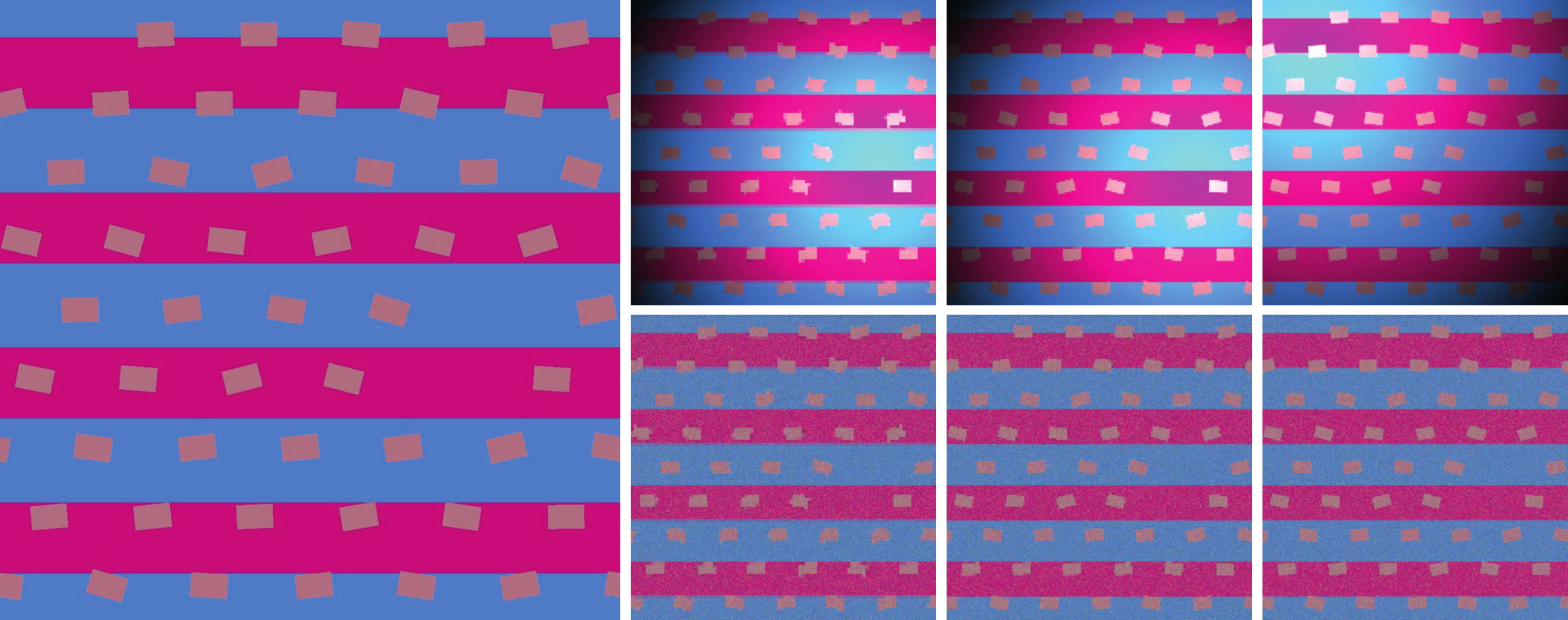}
\caption{\small \label{fig:exDist} Three examples of distortions. For each one the biggest image is the original pattern. On the right, the first row depicts the radial lighting effect while the second one the impulsive noise distortion. The column are organized from the 100x100 down-sampling to 300x300 down-sampling.}
\vspace{-0.5cm}
\end{figure}

Some examples of these images are shown in Fig.~\ref{fig:exDist}. It can be seen that distorted images simulate pictures that could be captured by users wishing to employ a retrieval application. The lighting effect simulates the flash of a camera while impulsive noise simulates general defects in the image acquisition process.

We consider each of the 6 variants as \emph{query set} and we test each one separately. 
Given a distorted image from the query set, the task is to retrieve the corresponding original one from the database set. The position of the correct match in the computed ranking is recorded. This process is repeated for every image in a query set.

To distance functions used for ranking is chosen according to the descriptor; for each descriptor we selected the best performing distance function between all of the ones available in the MATLAB software~\cite{MATLAB:2019}. More specifically, for the FV-CNN descriptor and our descriptor we employ the cosine distance while for the Tamura descriptor the cityblock distance function performs best.
\\
\\
Table~\ref{tab:RetriRes} shows the results of this experiment in all of the 6 variants previously described. In each case Texel-Att reaches the best results in terms of \emph{AUC: Area Under Curve} index related to CMC (Cumulative Matching Characteristics) curves shown in the plots in Fig~\ref{fig:Retr_plot}. We show only the first 200 positions for the CMC curve rank as we consider higher ranking positions less useful for a retrieval application (a user will rarely check results beyond 200 images).

%=================================CONCLUSIONS=====================================================
\section{Conclusion}\label{sec:conc}

This paper promotes to describe element-based textures by using attributes which focus on texels. Our framework, Texel-Att, can successfully describe and retrieve this type of patterns inside large databases even under simulated real-world factors such as poor resolution, noise and lighting conditions. The experiments show that we perform better in this task with our texel based attributes than by using state-of-the-art general texture descriptors, paving the way for retrieval applications in the fashion and textile domains where element-based textures are prominent.
\\
\\
\textbf{Acknowledgements:} This work has been partially supported by the project of the Italian Ministry of Education, 
Universities and Research (MIUR) "Dipartimenti di Eccellenza 2018-2022", and has been partially supported
by the POR FESR 2014-2020 Work Program (Action 1.1.4,
project No.10066183). We also thank Nicolò Lanza for assistance with Substance Designer software.
\bibliographystyle{splncs04}
\bibliography{egbib}

\end{document}